\documentclass[11pt]{article}
\PassOptionsToPackage{table}{xcolor}
\usepackage[preprint]{acl}
\usepackage{times}
\usepackage{latexsym}
\usepackage[T1]{fontenc}
\usepackage[utf8]{inputenc}
\usepackage{microtype}
\usepackage{graphicx}
\usepackage{booktabs}
\usepackage{array,tabularx}
\usepackage{amsmath,amssymb,amsthm}
\usepackage[table]{xcolor}
\usepackage{capt-of}
\usepackage{listings}
\usepackage{upquote}
\usepackage{hyperref}
\usepackage{cleveref} 

\lstdefinestyle{harnessfile}{
  basicstyle=\ttfamily\fontsize{8}{9.2}\selectfont,
  columns=fullflexible, keepspaces=true, showstringspaces=false,
  breaklines=true, breakatwhitespace=false, breakindent=1.5em,
  postbreak=\mbox{\textcolor{gray}{$\hookrightarrow$}\space},
  frame=single, framerule=0.3pt, rulecolor=\color{gray},
  aboveskip=3pt, belowskip=8pt, xleftmargin=0pt, xrightmargin=0pt,
  literate={—}{{---}}1,
}

\newtheorem{proposition}{Proposition}
\newtheorem{theorem}{Theorem}
\hypersetup{
  pdftitle={Forging More Tasks into Better Harnesses: Agentic Harness Evolution via Successive Scale Racing},
  pdfauthor={Jingzhe Liu, Jiliang Tang, Tong Zhao, Neil Shah, Clark Mingxuan Ju}
}

\lstdefinestyle{promptfile}{
  basicstyle=\small\ttfamily,
  columns=fullflexible, keepspaces=true, showstringspaces=false,
  breaklines=true, breakatwhitespace=false, breakindent=1em,
  frame=single, framerule=0.3pt, rulecolor=\color{gray},
  xleftmargin=0pt, xrightmargin=0pt,
}

\title{Forging More Tasks into Better Harnesses:\\ Agentic Harness Evolution via Successive Scale Racing}
\author{%
  \textbf{Jingzhe Liu}$^{1}$ \quad
  \textbf{Jiliang Tang}$^{1}$ \quad
  \textbf{Tong Zhao}$^{2}$ \\
  \textbf{Neil Shah}$^{2}$ \quad
  \textbf{Clark Mingxuan Ju}$^{2}$ \\
  $^{1}$Michigan State University \qquad $^{2}$Snap Inc.
}
\date{}

\begin{document}
\maketitle
\begin{abstract}
Agentic harness evolution optimizes an agent's instructions, tools, and control flow by using its task execution experiences as evolving evidence. 
While more evolution tasks should intuitively improve performance, our preliminary study reveals a paradoxical failure mode: the performance of existing methods degrades as the task pool expands. Our further analysis demonstrates that this degradation stems from suboptimal evidence selection during optimization, as different evidence combinations lead to substantially different evolution outcomes. To address this issue, we propose \textbf{Successive Scale Racing (SSR)}, a framework that adaptively searches for optimal evidence combinations in each evolution round. SSR constructs diverse evidence combinations based on the corresponding task quality and coverage, and efficiently races the resulting candidate harnesses under a small compute budget. Empirically, across three distinct benchmarks, SSR achieves state-of-the-art results in all six configurations. Notably, as the task pool grows from 40 to 200, SSR steadily improves the pass rate from 64.3\% to 68.3\%, while the baselines' performances all drop. Furthermore, its racing mechanism identifies near-optimal candidates while reducing validation executions by 60\%.
\end{abstract}

\section{Introduction}
\label{sec:introduction}
An agent harness governs a language model's interaction with its environment through instructions, tool interfaces, context management, and control flow \citep{ning2026codeharness,pan2026naturalharness}. By determining what the model observes, remembers, and can do \citep{packer2023memgpt}, it shapes how model capabilities are used on real-world tasks. Harness engineering has relied heavily on human design and implementation, such as manually refined interfaces \citep{yang2024sweagent} and coordination protocols \citep{hong2023metagpt,qian2023chatdev}. Recent work automates this process, evolving agent designs and workflows from execution experience while keeping model weights fixed \citep{hu2024adas,zhang2024aflow,zhang2026darwin}.

Current frameworks for agentic harness evolution commonly follow an execution--reflection--optimization paradigm \citep{lee2026metaharness,lin2026agentic,wang2026huxley,liu2026mendel}. The agent first solves assigned tasks, producing execution trajectories. An agent acting as the optimizer then distills lessons, which we refer to as evidence, from these trajectories, and uses them to guide improvements to the harness. This process repeats until a stopping criterion is met or the execution budget is exhausted.

\begin{figure}[!t]
\centering
\includegraphics[width=0.94\columnwidth]{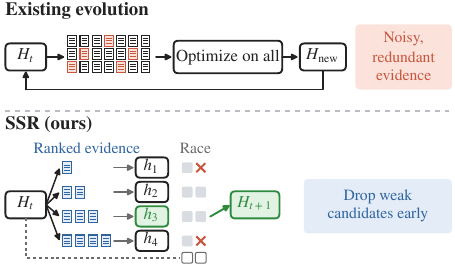}
\caption{Many existing methods evolve on large, fixed task pools, which can induce noisy or redundant signals. SSR searches over ordered evidence groups.}
\label{fig:intro-insight}
\end{figure}

In machine learning, expanding training task pools generally improves generalization and system capabilities \citep{kaplan2020scaling,hoffmann2022chinchilla}. However, whether agentic harness evolution similarly benefits from larger task pools remains underexplored.
To address this gap, we first evaluate existing state-of-the-art methods \citep{lin2026agentic,wang2026huxley,liu2026mendel,pan2026retrospective} by enlarging the set of training tasks used for evolution. Surprisingly, our preliminary study in Section~\ref{sec:preliminary} shows that larger task pools do not consistently yield better harnesses with existing methods. AHE \citep{lin2026agentic}, for example, initially improves but drops from 63.0\% at 80 tasks to 57.7\% at 200 tasks. Based on this observation, we pose our core research question: 
\begin{center}
\emph{How can we leverage experience from more tasks to evolve better harnesses?}
\end{center}

To provide further insights into this problem, we investigate how different trajectory selections impact harness updates. We find that a set of
tasks selected by a simple LLM judge can generate better harnesses than either an equally sized alternative subset or the much larger full task pool. 
This reveals a key insight: \emph{more task experiences, while providing more information, might also introduce noise or redundant signals}. Feeding all trajectories into the optimizer could blur the focus of individual modifications. Thus, harness performance depends not on the total volume of available evidence, but on selecting the appropriate combinations of evidence for each evolution step.

However, searching for effective evidence combinations presents two key challenges: (1) The space of possible trajectory combinations is combinatorial. Generating every candidate harness and thoroughly evaluating each on validation tasks is computationally prohibitive. (2) The ideal evidence scale is not static; it shifts as the parent harness evolves and handles tasks differently across iterations.

To address these challenges, we propose \textbf{Successive Scale Racing (SSR)} (Figure~\ref{fig:intro-insight}), a framework that adaptively searches over evidence combinations under a fixed execution budget. SSR addresses the two challenges in the following ways: (1) Instead of enumerating combinatorial subsets, SSR introduces Representative Evidence Ranking (RER) to order task evidence by quality and coverage. It generates candidates from a selected set of evidence combinations based on the rankings. To minimize validation costs, SSR employs a cumulative performance racing mechanism that evaluates candidates on shared task batches and progressively prunes weaker candidates early. (2) In every evolution round, SSR refreshes execution trajectories based on the incumbent harness and dynamically re-ranks evidence subsets, allowing the winning evidence combination to adaptively shift as the harness evolves.

We evaluate SSR on SWE-Bench Pro Verified, DeepSWE, and GAIA-2 \citep{zheng2026swebenchproverified,huang2026deepswe,froger2026gaia2} across two execution setups. SSR achieves top performance and exhibits positive scaling, raising accuracy from 64.3\% to 68.3\% as the task pool expands from 40 to 200 tasks.
Our main contributions are threefold: (1) We identify noisy and redundant evidence signals as a primary cause of performance degradation when scaling task pools in harness evolution. (2) We propose SSR, which effectively searches over evidence scales using quality-coverage ranking and performance racing. (3) Across three benchmarks under two execution setups, SSR leads in all six settings and benefits from larger task pools, achieving higher efficiency and performance.

\section{Related Work}
\label{sec:related-work}

\subsection{Agent Harness Engineering}
Harness design encompasses reasoning--action loops \citep{yao2023react}, multi-agent conversation \citep{wu2023autogen}, sandboxed execution \citep{wang2024openhands}, and prescribed repair workflows \citep{xia2024agentless}. Automated optimization spans instruction search \citep{zhou2023ape,yang2023opro}, feedback-guided prompt revision \citep{pryzant2023protegi,agrawal2025gepa}, pipeline compilation \citep{khattab2023dspy}, and textual feedback propagation \citep{yuksekgonul2024textgrad}. Beyond prompts, GPTSwarm optimizes computational graphs \citep{zhuge2024gptswarm}, AgentSquare evolves and recombines modules \citep{shang2024agentsquare}, and AgentOptimizer updates tool functions \citep{zhang2024agentoptimizer}.

\begin{figure*}[!t]
  \begin{minipage}[t]{0.48\textwidth}
    \centering
    \includegraphics[width=0.96\linewidth]{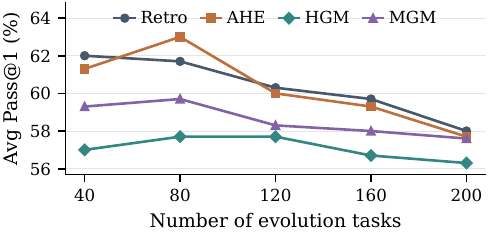}
    \caption{Existing methods' performance change with more evolution tasks. Larger task pools do not consistently improve final harness performance.}
    \label{fig:prelim-pool}
  \end{minipage}\hfill
  \begin{minipage}[t]{0.48\textwidth}
    \centering
    \includegraphics[width=0.96\linewidth]{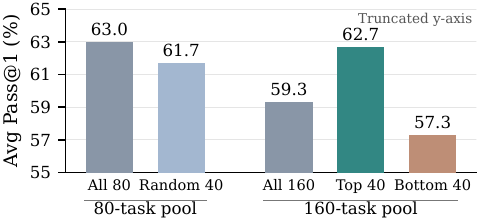}
    \caption{Comparisons of harness performance on different task pools. A smaller, selected task pool tends to induce better harnesses than a large, random task pool.}
    \label{fig:prelim-evidence}
  \end{minipage}
\end{figure*}

\subsection{Self-Evolving Agents and Harnesses}
Experience can improve outputs through self-feedback \citep{madaan2023selfrefine}, later attempts through verbal reflection \citep{shinn2023reflexion}, and future tasks through reusable skills \citep{wang2023voyager}, extracted insights \citep{zhao2023expel}, contextual guidelines \citep{fu2024autoguide}, instruction manuals \citep{chen2024automanual}, reasoning memory \citep{ouyang2026reasoningbank}, or evolving playbooks \citep{zhang2025ace}. Code-level evolution includes SICA's self-edits \citep{robeyns2025sica}, DGM's archive \citep{zhang2026darwin}, HGM's descendant-based search \citep{wang2026huxley}, and MGM's cross-task and cross-lineage comparisons \citep{liu2026mendel}. AHE structures trajectory evidence \citep{lin2026agentic}, while Meta-Harness exposes historical code, scores, and traces \citep{lee2026metaharness}. Retro selects a challenging, diverse coreset and uses self-preference for retrospective optimization \citep{pan2026retrospective}. 

\section{Preliminary Study}
\label{sec:preliminary}

In this preliminary study, we investigate how task-pool size and evidence use affect harness evolution. To start with, we conduct experiments on SWE-Pro-V with the Codex setting, as in the analyses of Section~\ref{sec:experiments}. We report final Avg Pass@1~(\%) averaged over three independent runs. Section~\ref{sec:exp-evidence} extends the evidence comparison to DeepSWE.

\subsection{Effect of Task-Pool Size}
\label{subsec:effect_task_pool}
We vary the training task pool from 40 to 200 tasks across Retro, AHE, HGM, and MGM. Contrary to intuition, none of these methods consistently benefits from larger pools (Figure~\ref{fig:prelim-pool}). For instance, AHE peaks at 63.0\% accuracy with 80 tasks, but degrades to 57.7\% with 200 tasks. 

\paragraph{\textbf{Observation 1.}}\emph{Expanding the task pool does not necessarily produce a better harness under existing evolution strategies, and can even degrade performance.}

\subsection{Effect of Evidence Configuration}
\label{subsec:effect_evidence_config}
To understand this degradation, we analyze how different evidence selections impact optimization starting from the same harness (Figure~\ref{fig:prelim-evidence}). An LLM judge ranks experience records from the 160-task pool into \textsc{Top 40} and \textsc{Bottom 40} subsets. \textsc{Top 40} outperforms \textsc{All 160} by 3.4\% and \textsc{Bottom 40} by 5.4\%, showing that evidence quality and scale matter significantly. Conversely, a randomly sampled 40-record subset from the 80-task pool trails \textsc{All 80} by 1.3\%, confirming that unguided scale reduction is unhelpful.

\paragraph{\textbf{Observation 2.}} \emph{Evolution outcomes depend on the evidence combinations (from task trajectories) the optimizer uses; a well-selected subset can outperform full-pool evidence.}
\begin{figure*}[t]
  \centering
  \input{figures/method/method_figure}
\end{figure*}

\section{Successive Scale Racing}
\label{sec:method}

In this section, we introduce \textbf{Successive Scale Racing (SSR)}. As illustrated in Figure~\ref{fig:ssr-method}, our framework operates in four key stages: (1) Representative Evidence Ranking (RER), which orders experience records by quality and coverage; (2) Multiscale Candidate Generation, which independently modifies a shared parent harness using nested evidence prefixes generated from the ranking; (3) Candidate Performance Racing, which progressively prunes weaker candidates using cumulative scores on shared validation batches; and (4) Recursive Evolution, which updates the parent harness and refreshes execution evidence for subsequent rounds.

\paragraph{Notation.}
Let $H_t$ denote the incumbent harness at round $t$, and let $\mathcal{T}$ and $\mathcal{V}$ be disjoint training and validation task sets. Training tasks supply optimization evidence; validation tasks are held out from that evidence and contribute to candidate selection. $N=|\mathcal{T}|$ denotes the available evidence-pool size, $k$ the number of task records supplied to one candidate, and $M$ the number of candidates in a round.

\subsection{Representative Evidence Ranking}
\label{sec:rer}

An effective evidence set must offer actionable and transferable insights without overwhelming the optimizer with redundant or low-value execution logs. To achieve this balance, Representative Evidence Ranking (RER) establishes a global ordering of experience records that jointly optimizes individual evidence quality and collective pool coverage.

\paragraph{Experience characterization.}
For a given parent harness $H_t$, we execute it on each training task $i \in \{1, \ldots, N\}$ to collect one execution trajectory $\tau_i$. An LLM judge distills each trajectory $\tau_i$ along with its feedback into a structured experience record $e_i$, which comprises an executive summary and concrete harness improvement suggestions. 

Formally, for each record $e_i$, the judge outputs three normalized metrics in $[0,1]$: experience importance $w_i$, actionability $a_i$, and confidence $u_i$. Here, $w_i$ quantifies the generalizability of the experience beyond its originating task; $a_i$ measures the concreteness of the proposed harness modifications; and $u_i$ reflects the judge's assessment reliability. We define the intrinsic record quality as $q_i = a_i u_i$, ensuring a high score only when a concrete modification is proposed with high confidence. The importance weight $w_i$ is utilized separately to parameterize coverage over the evidence pool (Eq.~\ref{eq:ssr-coverage}).

Because trajectories from different tasks can reveal overlapping failure modes or identical improvement directions, we explicitly model pairwise evidence redundancy. Specifically, each record $e_i$ is mapped to a feature embedding $z_i$ by encoding its summary and extracted insights. We quantify the nonnegative pairwise similarity between records $e_i$ and $e_j$ as:
\begin{equation}
 s_{ij}=\max\{0,\cos(z_i,z_j)\}.
 \label{eq:ssr-similarity}
\end{equation}
This similarity metric $s_{ij}$ directly feeds into the coverage objective to measure the extent to which a selected evidence subset represents the complete trajectory pool.

\paragraph{Coverage with diminishing returns.}
For a selected set $Y\subseteq\{1,\ldots,N\}$ and each record $j$ in the pool, let $C_j(Y)$ be the quality-weighted similarity of the selected records to $e_j$, that is, how well $Y$ covers it. The objective $F(Y)$ sums the log-coverage of all records, weighted by importance:
\begin{align}
 C_j(Y)&=\sum_{i\in Y}q_i s_{ij}, \nonumber\\
 F(Y)&=\sum_{j=1}^{N}w_j\log\bigl(1+C_j(Y)\bigr).
 \label{eq:ssr-coverage}
\end{align}
The logarithm gives diminishing returns: a record similar to already-covered ones raises $F$ by less, leaving room for other important experiences. $F$ is monotone and submodular (Proposition~\ref{prop:ssr-rer}), which the greedy ordering below exploits.

Starting from $Y_0=\varnothing$, RER repeatedly selects the record with the largest marginal coverage gain:
\begin{align}
 \pi_r&\in\arg\max_{i\notin Y_{r-1}}
   \bigl[F(Y_{r-1}\cup\{i\})-F(Y_{r-1})\bigr],\nonumber\\
 Y_r&=Y_{r-1}\cup\{\pi_r\}.
 \label{eq:ssr-ranking}
\end{align}
The resulting permutation $\pi=(\pi_1,\ldots,\pi_N)$ defines every evidence prefix. Marginal gains are recomputed after each selection; this is not a static sort by individual quality.

\subsection{Multiscale Candidate Generation}
\label{sec:ssr-proposals}

Because the optimal evidence scale is unknown \emph{a priori} and shifts as evolution proceeds (Appendix~\ref{sec:exp-scale-dynamics}), SSR treats evidence scale as a per-round search variable. Given a set of predefined scale budgets $\mathcal{S}=\{k_1,\ldots,k_M\}$ ($1\leq k_1<\cdots<k_M=N$), we construct nested evidence prefixes from the ranking $\pi$:
\begin{equation}
 D_{t,k}=\{e_{\pi_1},\ldots,e_{\pi_k}\}, \qquad k\in\mathcal{S}.
 \label{eq:ssr-prefix}
\end{equation}
Using nested prefixes rather than disjoint subsets offers distinct advantages. First, every prefix $D_{t,k}$ provides near-optimal coverage at its length, guaranteeing a $(1-1/e)$ approximation factor (Proposition~\ref{prop:ssr-rer}; \citealp{feige1998threshold}). Second, nested structures isolate scale as the sole variable across candidates, preventing confounding between subset composition and evidence volume (Section~\ref{sec:exp-evidence}). Third, the shared evidence core facilitates effective validation batching during performance racing (Section~\ref{sec:ssr-racing}).

For each scale $k \in \mathcal{S}$, the optimizer independently applies edits to the same frozen parent harness $H_t$:
\begin{equation}
 H_t^{(k)}=\operatorname{Optimize}(H_t,D_{t,k}).
 \label{eq:ssr-proposal}
\end{equation}
The optimizer receives structured experience summaries and improvement suggestions from $D_{t,k}$, with optional access to raw trajectories. All candidate harnesses $H_t^{(k)}$ are generated under identical optimizer prompts and permissions, serving as temporary hypotheses to be evaluated in the subsequent racing phase.

\subsection{Candidate Performance Racing}
\label{sec:ssr-racing}

Evaluating every proposed harness on every task would spend substantial execution budget on weak candidates. This is a fixed-budget best-arm identification problem \citep{audibert2010best,karnin2013almost}: each harness is an arm, each task execution is a pull, and the goal is to return the best harness within the budget. SSR follows the successive-halving schedule \citep{karnin2013almost,jamieson2016nonstochastic} with a constant elimination quota, and protects the parent as an incumbent. SSR introduces new tasks over $K$ stages and directs later evaluations to surviving candidates. The parent remains active throughout as a common reference and is never included in the child-elimination quota. Evaluation continues across stages even when only one child remains.

\paragraph{Shared evaluation batches.}
At stage $r$, every surviving child and the parent execute the same new batch
\begin{equation}
 B_r=B_r^{\mathrm{val}}\cup B_r^{\mathrm{succ}}
                  \cup B_r^{\mathrm{fail}}.
 \label{eq:ssr-batch}
\end{equation}
The validation component is sampled randomly without replacement from $\mathcal{V}$. The other components contain training tasks on which the parent succeeded or failed, respectively, with these labels fixed within the round. They favor high-ranked tasks shared by the surviving candidates' evidence prefixes, supplementing validation with checks of retained behavior and repair on familiar tasks. The prescribed numbers of new tasks from each source are constant across stages, subject to task availability; detailed sampling rules are deferred to Appendix~\ref{app:ssr-implementation}.

Let $P_r=\bigcup_{s=1}^{r}B_s$ be the cumulative task set. Only new tasks are executed at each stage; earlier scores remain cached. With $R$ executions per harness--task pair and reward $y_{h,x,\ell}\in[0,1]$, the cumulative score is
\begin{equation}
 \widehat\mu_{h,r}
 =\frac{1}{|P_r|}\sum_{x\in P_r}
       \left(\frac{1}{R}\sum_{\ell=1}^{R}y_{h,x,\ell}\right).
 \label{eq:ssr-score}
\end{equation}
Tasks from all three sources receive equal weight. The selection score therefore measures performance on this mixed evaluation set; it is not a validation-only estimate. All surviving harnesses have results on the same cumulative tasks, enabling direct parent--child and child--child comparisons.

\paragraph{Elimination and selection.}
After each intermediate stage, SSR ranks surviving children by $\widehat\mu_{h,r}$ and removes a predetermined number of the lowest-scoring candidates. The quota is based on the initial number $M$, approximately $M/K$ per stage, rather than a fraction of the current survivor count. Integer quotas retain at least one child for final selection. Eliminated candidates receive no further executions. Ties favor the candidate with fewer added and deleted lines relative to $H_t$.

Let $r_\star$ be the last completed stage and $h^\star$ the highest-scoring surviving child, using the same tie-break rule. SSR updates the incumbent only for a strict empirical improvement:
\begin{equation}
 H_{t+1}=\begin{cases}
 h^\star,&\widehat\mu_{h^\star,r_\star}>
                         \widehat\mu_{H_t,r_\star},\\
 H_t,&\text{otherwise}.
 \end{cases}
 \label{eq:ssr-update}
\end{equation}
\subsection{Recursive Evolution}
\label{sec:ssr-evolution}

After an accepted update, SSR reruns the new parent on all training tasks and rebuilds the experience records, outcome labels, and RER ordering. If the parent is retained, it reuses these records and the ordering while generating fresh candidates. Racing scores are accumulated within each round. Training-trajectory acquisition and all parent and child evaluation executions count toward the execution budget $B$. Evolution ends when the execution budget is exhausted or no strict improvement has occurred for $P$ consecutive rounds, and returns the current incumbent. This maintains a single evolving harness while repeatedly exploring alternative evidence combinations.

\section{Theoretical Analysis}
\label{sec:ssr-theory}
We give single-round guarantees for evidence coverage and candidate selection; full statements and proofs are in Appendix~\ref{app:ssr-proofs}.

\paragraph{Proposition~\ref{prop:ssr-rer} (RER prefixes).}
\textit{For a fixed evidence pool with nonnegative weights and similarities, $F$ is normalized, monotone, and submodular. Every length-$k$ prefix $D_k$ of its exact greedy ordering satisfies}
\begin{equation*}
 F(D_k)\geq(1-1/e)\max_{|D|\leq k}F(D).
\end{equation*}
One ranking thus provides coverage guarantees at every candidate scale \citep{nemhauser1978analysis}, supporting comparisons between focused and broad evidence. The guarantee concerns $F$, not execution reward.

\paragraph{Theorem~\ref{thm:ssr-racing} (Single-round racing).}
\textit{Fix the parent and its children, forming $\mathcal C$, and let $\mu(h)$ be the expected reward over the finite validation pool. Assume a fresh uniform validation permutation, fixed stage sizes, predictable training-task choices, and fresh rollouts in $[0,1]$ with stationary conditional means. With probability at least $1-\delta$, at every completed stopping stage $\nu\leq K$,}
\begin{align*}
 \max_{h\in\mathcal C}\mu(h)-\mu(\widehat H)
 &\leq 2\sum_{r=1}^{\nu}\varepsilon_r,\\
 \mu(H_t)-\mu(\widehat H)&\leq2\varepsilon_\nu.
\end{align*}
Here $\varepsilon_r$ is the mixed-score error radius in Eq.~\eqref{eq:ssr-race-radius}. Its three components are rollout noise, validation-task sampling error, and training--validation discrepancy weighted by the training fraction.

The first bound limits the cost of choosing among evidence scales, even if the best child is eliminated: each elimination can lose at most $2\varepsilon_r$ on the joint error-control event. The second bounds regression relative to the protected parent using only the final-stage error. More rollouts reduce execution noise but cannot remove training--validation discrepancy. These are conditional, single-round guarantees on a finite pool; they establish neither population generalization nor monotonic improvement across evolution rounds.

\begingroup
\providecolor{SSRBlue}{HTML}{196C94}
\providecolor{SSRHeader}{HTML}{EDF1F5}
\providecolor{SSRHighlight}{HTML}{DAEDF5}
\providecolor{SSRGain}{HTML}{F0F7FA}
\begin{table*}[!t]
\centering
\small
\setlength{\tabcolsep}{7pt}
\renewcommand{\arraystretch}{1.07}
\begin{tabular}{lrrrrrr}
\toprule
\rowcolor{SSRHeader}
& \multicolumn{3}{c}{\textbf{Codex: GPT 5.5 high}} & \multicolumn{3}{c}{\textbf{Claude Code: Opus 4.8 high}} \\
\rowcolor{SSRHeader}
Method & SWE-Pro-V & DeepSWE & GAIA-2 & SWE-Pro-V & DeepSWE & GAIA-2 \\
\midrule
Initial harness & 57.7 & 59.3 & 29.0 & 56.3 & 56.7 & 26.0 \\
Meta-Harness & 59.7 & 60.7 & 31.7 & 60.0 & 58.0 & 29.3 \\
HGM & 58.7 & 62.7 & 32.3 & 60.3 & 58.7 & 29.3 \\
MGM & 60.0 & 63.3 & 32.3 & 61.0 & 60.7 & 29.7 \\
AHE & 60.7 & 65.3 & 36.0 & \underline{62.0} & 63.3 & 33.7 \\
Retro & \underline{62.7} & \underline{66.7} & \underline{36.3} & 61.3 & \underline{64.0} & \underline{34.3} \\
\midrule
\rowcolor{SSRHighlight}
\textbf{SSR (ours)} & \textbf{65.7} & \textbf{68.0} & \textbf{38.0} & \textbf{63.3} & \textbf{65.3} & \textbf{36.0} \\
\bottomrule
\end{tabular}
\captionsetup{hypcap=false}
\captionof{table}{Final harness performance (Avg Pass@1, \%). SWE-Pro-V denotes SWE-Bench Pro Verified. Each entry is the mean over three independent evolution runs; standard deviations are in Appendix~\ref{app:main-results-variance}. Bold marks the best score; underlining marks the strongest baseline.}
\label{tab:exp-main}

\end{table*}
\section{Experiments}
\label{sec:experiments}

We conduct extensive experiments on SSR to answer the following questions:
\begin{list}{}{\setlength{\leftmargin}{2.1em}\setlength{\labelwidth}{1.8em}\setlength{\labelsep}{0.3em}\setlength{\itemsep}{1pt}\setlength{\parsep}{0pt}\setlength{\topsep}{3pt}}
\item[\textbf{Q1.}] How does SSR compare with existing harness evolution methods (Section~\ref{sec:exp-overall})?
\item[\textbf{Q2.}] Does SSR benefit from more task experience and search resources (Section~\ref{sec:exp-experience})?
\item[\textbf{Q3.}] How do evidence selection strategies affect performance (Section~\ref{sec:exp-evidence})?
\item[\textbf{Q4.}] How does racing trade validation cost for selection quality (Section~\ref{sec:exp-racing})?
\end{list}

\subsection{Experimental Setup}
\label{sec:exp-setup}
\paragraph{Benchmarks and models.}
We use SWE-Bench Pro Verified (SWE-Pro-V; \citealp{zheng2026swebenchproverified}), DeepSWE \citep{huang2026deepswe}, and GAIA-2 \citep{froger2026gaia2}, with disjoint train/validation/test splits of 80/20/100, 50/13/50, and 80/20/100 tasks. Training trajectories supply evidence for candidate generation, validation tasks support selection, and test tasks assess the resulting harness. For GAIA-2, we adopt Retro's 200-scenario mini configuration from the official validation split, its four-message-per-turn cap, and no optional judge relaxations \citep{pan2026retrospective}. Our execution settings are Codex with GPT 5.5 high and Claude Code with Opus 4.8 high. The optimizer matches the executor's model and reasoning configuration.

\paragraph{Baseline methods.}
We compare SSR with the initial harness, Meta-Harness \citep{lee2026metaharness}, HGM \citep{wang2026huxley}, MGM \citep{liu2026mendel}, AHE \citep{lin2026agentic}, and Retro \citep{pan2026retrospective}. Within each platform, methods start with the same empty additional harness while retaining the platform's default tools and runtime. Baselines follow their original evolution procedures. SSR generates one candidate from each RER prefix of 25\%, 50\%, 75\%, and 100\%, using a common parent. 

\paragraph{Experimental control.}
All subsequent analytical experiments are conducted using Codex and Claude Code. When isolating specific components (e.g., fixed-strategy vs. ordering variants), we keep the candidate count, optimizer, racing mechanism, and execution cap identical to ensure fair comparison.

\paragraph{Evaluation protocol.}
For our main results (Table~\ref{tab:exp-main}), each end-to-end condition is evolved across three independent runs. We evaluate the resulting final harness on test tasks using Average Pass@1 across three trials, reporting the mean performance over the three runs (with standard deviations detailed in Appendix~\ref{app:main-results-variance}). Ablation studies and controlled analyses use a single evolution run per condition under the same evaluation protocol, whereas the racing audit relies on three selection replays over a fixed trajectory matrix. All reported performance differences represent absolute percentage point changes. Full experimental configurations are provided in Appendix~\ref{app:experimental-configuration}.

\subsection{Overall Performance}
\label{sec:exp-overall}

As reported in Table~\ref{tab:exp-main}, SSR achieves state-of-the-art performance across all six benchmark--execution configurations. Under Codex (GPT 5.5 high), SSR outperforms the strongest baselines by 3.0\%, 1.3\%, and 1.7\% on SWE-Bench Pro Verified (65.7\%), DeepSWE (68.0\%), and GAIA-2 (38.0\%), respectively. Under Claude Code (Opus 4.8 high), SSR maintains consistent superiority, exceeding the top-performing baselines by 1.3\% on SWE-Bench Pro Verified (63.3\%), 1.3\% on DeepSWE (65.3\%), and 1.7\% on GAIA-2 (36.0\%). These results confirm that SSR provides robust performance gains across diverse task domains and models.

\begin{figure*}[!t]
\centering
\includegraphics[width=0.96\textwidth]{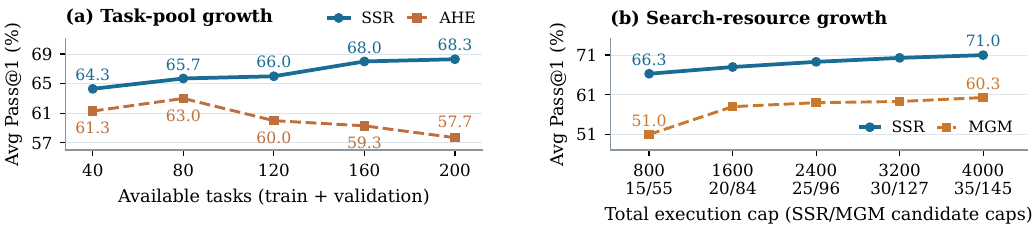}
\caption{Scalability of harness evolution on SWE-Bench Pro Verified.
\textbf{(a) Task-pool growth:} Evaluation performance when expanding available tasks from 40 to 200 with proportionally scaled execution caps. 
\textbf{(b) Search-resource growth:} Performance trajectory when increasing total execution caps under a fixed task pool.}
\label{fig:exp-growth}
\par\vspace{\floatsep}
\includegraphics[width=0.96\textwidth]{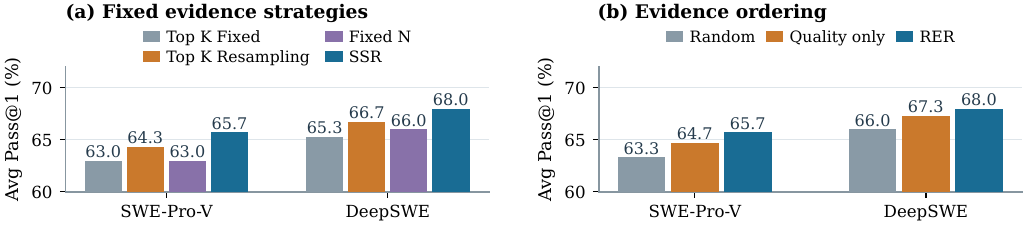}
\caption{Ablation studies on evidence strategies and ordering mechanisms.
\textbf{(a) Fixed evidence strategies:} Performance comparison of SSR against fixed-evidence baselines under identical candidate counts and racing protocols. 
\textbf{(b) Evidence ordering:} Performance comparison of RER against Random and Quality-only ordering under the same nested scale construction.}
\label{fig:exp-evidence}
\end{figure*}

\subsection{Effects of Task Experience and Search Resources}
\label{sec:exp-experience}

\paragraph{Scaling task pools.}
We first evaluate how harness evolution scales as the available task pool expands from 40 to 200 tasks, with execution budgets scaled proportionally. As shown in Figure~\ref{fig:exp-growth}a, SSR exhibits monotonic performance gains, steadily raising accuracy from 64.3\% to 68.3\%. In stark contrast, AHE suffers from severe degradation when task volume grows, peaking at 63.0\% at 80 tasks before plummeting to 57.7\% at 200 tasks. This confirms that SSR successfully converts larger task pools into continuous harness improvements.

\paragraph{Scaling search resources.}
Next, holding the task pool fixed, we scale the total execution budget from 800 to 4{,}000 while expanding candidate-generation caps (from 15 to 35 for SSR, and from 55 to 145 for MGM). As illustrated in Figure~\ref{fig:exp-growth}b, increased compute budget consistently benefits both methods, with SSR improving from 66.3\% to 71.0\% and maintaining a decisive lead over MGM (51.0\% to 60.3\%) at every checkpoint. These results demonstrate that SSR can effectively leverage additional compute to explore and identify superior harnesses.

\subsection{Effects of Evidence Selection Strategies}
\label{sec:exp-evidence}
\paragraph{Comparison with fixed evidence strategies.}
To isolate the benefit of multiscale evidence, we compare SSR against fixed-evidence variants across four candidate generators: \textsc{Top K Fixed} (supplying the same top 25\% of the evidence), \textsc{Top K Resampling} (independent 25\% subsets), and \textsc{Fixed N} (full evidence set). As shown in Figure~\ref{fig:exp-evidence}a, relying on repeated sampling from identical evidence (\textsc{Top K Fixed}) or feeding full evidence (\textsc{Fixed N}) yields suboptimal performance on both benchmarks. While varying evidence composition (\textsc{Top K Resampling}) improves outcomes, SSR's multiscale strategy provides a further gain of 1.4\% on SWE-Pro-V (65.7\% vs.\ 64.3\%) and 1.3\% on DeepSWE (68.0\% vs.\ 66.7\%), demonstrating the distinct necessity of searching across evidence scales.

\paragraph{RER ordering ablation.}
We evaluate the necessity of RER by comparing it with \textsc{Random} ordering and a \textsc{Quality only} baseline under the same nested-prefix construction. As illustrated in Figure~\ref{fig:exp-evidence}b, ordering evidence by quality alone improves over random ranking on both benchmarks. Incorporating representative coverage via RER yields an additional performance boost of 1.0\% on SWE-Pro-V (65.7\% vs.\ 64.7\%) and 0.7\% on DeepSWE (68.0\% vs.\ 67.3\%), confirming that joint optimization of quality and coverage provides superior evidence guidance.

\begin{figure}[!t]
\centering
\includegraphics[width=0.94\columnwidth]{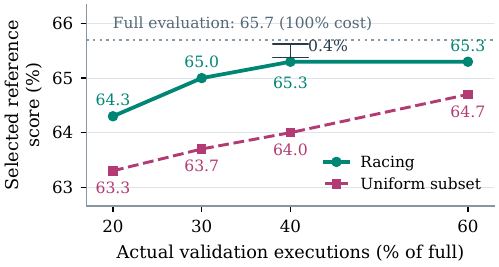}
\caption{Racing quality versus validation cost, averaged over three task-order replays of a fixed pool. Racing uses 40\% of the full-evaluation cost to achieve a score within 0.4\% of the best full-evaluation score.}
\label{fig:exp-racing}
\end{figure}

\subsection{Racing Cost--Quality Audit}
\label{sec:exp-racing}

To evaluate the efficiency of our selection mechanism, we isolate 12 candidate harnesses (three per evidence scale) generated from a single parent. We construct a 160-task audit matrix, disjoint from candidate-generation and test sets, and compare Racing against a Uniform Subset baseline that distributes evaluation budgets equally across all candidates.

As shown in Figure~\ref{fig:exp-racing}, Racing consistently outperforms uniform allocation across all validation budget levels (20\%--60\%). Notably, at just 40\% of the full validation cost, Racing achieves a 65.3\% score---within 0.4\% of the full-matrix upper bound (65.7\%)---while saving 60\% of validation executions. Increasing the budget to 60\% yields no further performance gains. These results confirm that SSR's adaptive racing mechanism achieves near-optimal candidate selection with a substantially reduced evaluation budget.
\endgroup

\section{Conclusion}
\label{sec:conclusion}
We propose Successive Scale Racing (SSR) to address performance degradation when scaling task pools in harness evolution. SSR ranks task experience by quality and coverage, generates multiscale candidate harnesses, and efficiently races them under a fixed compute budget. Across three benchmarks and two execution setups, SSR achieves top performance in all six settings and enables positive task-pool scaling. Furthermore, its racing mechanism identifies near-optimal candidates while cutting validation executions by 60\%, confirming that evidence-driven candidate diversity and execution-based selection provide a robust paradigm for agent self-improvement.
\clearpage
\section*{Limitations}
\label{sec:limitations}

While Successive Scale Racing (SSR) demonstrates promising results in scaling agentic harness evolution, several limitations warrant further investigation. First, the performance degradation observed when expanding training task pools is a multifaceted phenomenon. While our study focuses on noisy, conflicting, and redundant evidence as a primary underlying driver, other systemic factors may also contribute to this failure mode. Second, searching over evidence scales via SSR represents an initial step toward mitigating this scaling bottleneck. Alternative paradigms, such as adaptive evidence filtering via dynamic context pruning, fine-grained trajectory credit assignment, or reinforcement learning-based optimizer tuning, remain promising unexplored avenues for addressing noisy feedback during evolution. Third, our empirical evaluation, though conducted across multiple benchmarks and execution setups, is bounded by the specific choice of foundation models used as task agents and optimizers. Further validation is needed to determine how well these findings generalize to a broader spectrum of model architectures, such as open-weight models of varying scales or domain-specialized language models.

\section*{Ethical Considerations}

Improving agent harnesses can increase both the benefits of automated task completion and the potential harms of erroneous or misused tool actions. Our benchmark evaluations do not establish safety in unrestricted deployment. Practical use should retain execution isolation, restricted tool permissions, and human oversight for consequential actions. Task experience may also contain sensitive information or biases that can propagate into evolved harnesses; trajectories and generated artifacts should therefore be reviewed before sharing or deployment. The use and redistribution of third-party datasets, models, and code should respect their respective licenses and terms.
\bibliography{bibliography/references}

\clearpage
\appendix
\numberwithin{equation}{section}
\renewcommand{\theequation}{\thesection\arabic{equation}}
\renewcommand{\theHequation}{appendix.\thesection.\arabic{equation}}
\section{Proofs and Technical Conditions}
\label{app:ssr-proofs}
This appendix proves Proposition~\ref{prop:ssr-rer} and Theorem~\ref{thm:ssr-racing} for the confirmed SSR procedure. The racing proof uses ranking-based fixed quotas, mixed equal-task-weight scores, a protected parent, and selection after a fully completed stage. No confidence threshold is added to the algorithm.

\subsection{Formal Guarantees}
\label{app:ssr-statements}
\paragraph{Coverage across evidence scales.}
Fix one evidence pool and its nonnegative $w_j,q_i,s_{ij}$.
\begin{proposition}[RER prefixes]
\label{prop:ssr-rer}
The RER objective $F$ is normalized, monotone, and submodular. Every length-$k$ prefix $D_k$ of its exact marginal-gain greedy ordering satisfies
\begin{equation}
 F(D_k)\geq(1-1/e)\max_{|D|\leq k}F(D).
 \label{eq:ssr-prefix-guarantee}
\end{equation}
\end{proposition}
This applies the classical greedy guarantee of \citet{nemhauser1978analysis}. One ordering therefore supplies nested inputs with a coverage guarantee at every scale. The guarantee concerns $F$, not execution reward.

\paragraph{Selection under mixed evaluation.}
Condition on a fixed parent and its $M$ generated children, forming $\mathcal C$ with $L=M+1$ members. Let $\mu(h)$ be the expected reward averaged over the finite validation pool $\mathcal V$. At stage $r$, write $n_r=|P_r|$, $v_r=|P_r\cap\mathcal V|\geq1$, and $\alpha_r=v_r/n_r$. Let $\beta_r$ be the largest absolute difference, over $h\in\mathcal C$, between its mean expected reward on the selected training tasks and $\mu(h)$ (zero if none are selected). With $R$ rollouts per harness--task pair, define
\begin{align}
 c_\delta&=\log(4LK/\delta),\nonumber\\
 \varepsilon_r&=
 \sqrt{\frac{c_\delta}{2Rn_r}}
 +\alpha_r\sqrt{\frac{c_\delta}{2v_r}}
 +(1-\alpha_r)\beta_r.
 \label{eq:ssr-race-radius}
\end{align}
\begin{theorem}[Single-round racing]
\label{thm:ssr-racing}
Assume validation tasks follow a fresh uniform random permutation, stage sizes are fixed subject to availability, and fresh bounded rollouts have stationary conditional means. Training-task choices may depend on earlier stages, but precede their new outcomes. Under SSR's fixed-quota elimination and empirical parent-update rule, with probability at least $1-\delta$, the output $\widehat H$ satisfies, simultaneously at every completed stopping stage $\nu\leq K$,
\begin{align}
 \max_{h\in\mathcal C}\mu(h)-\mu(\widehat H)
 &\leq 2\sum_{r=1}^{\nu}\varepsilon_r,\label{eq:ssr-selection-guarantee}\\
 \mu(H_t)-\mu(\widehat H)&\leq2\varepsilon_\nu.
 \label{eq:ssr-parent-guarantee}
\end{align}
\end{theorem}
Theorem~\ref{thm:ssr-racing} is a simple-regret bound for the identification problem of Section~\ref{sec:ssr-racing}. Unlike standard successive-halving analyses \citep{karnin2013almost,jamieson2016nonstochastic}, which require the best arm to survive every stage, it allows mistaken eliminations and accumulates their possible losses, and it accounts for finite-pool validation sampling and adaptively chosen training tasks. Because the candidate set contains one candidate per scale, Eq.~\eqref{eq:ssr-selection-guarantee} also bounds the cost of not knowing the right scale: for every $k\in\mathcal{S}$, $\mu(\widehat H)\geq\mu(H_t^{(k)})-2\sum_{r\leq\nu}\varepsilon_r$, so SSR is never much worse than committing to any single scale, and gains whenever another scale's candidate is better. The three terms in $\varepsilon_r$ separate rollout noise, validation-task sampling, and training-to-validation discrepancy. More rollouts reduce the first term, not the discrepancy. Keeping the parent limits final regression by the last-stage error. These are conditional finite-pool guarantees, not claims of population generalization or monotonic multi-round improvement.

\subsection{Proof of the RER Prefix Guarantee}
\label{app:ssr-rer-proof}
Identify records with their indices in $E=\{1,\ldots,N\}$. The method uses
\begin{align}
 C_j(Y)&=\sum_{i\in Y}q_i s_{ij},\nonumber\\
 F(Y)&=\sum_{j=1}^{N}w_j\log(1+C_j(Y)).
\end{align}
The judge scores and clipped similarities are fixed within the ordering. Since they are nonnegative, $F(\varnothing)=0$ and adding an element cannot decrease $F$. For $A\subseteq B$ and $i\notin B$, its marginal gain is
\begin{equation}
 \Delta(i\mid Y)=\sum_jw_j\log\left(1+
       \frac{q_i s_{ij}}{1+C_j(Y)}\right).
\end{equation}
Because $C_j(A)\leq C_j(B)$, each summand is at least as large for $A$ as for $B$. Hence $\Delta(i\mid A)\geq\Delta(i\mid B)$: $F$ is submodular.

Fix $k$ and an optimal set $O_k$ with $|O_k|\leq k$. Let $G_r$ be the first $r$ greedy indices. Monotonicity and diminishing returns give
\begin{align}
 F(O_k)-F(G_r)
 &\leq\sum_{i\in O_k\setminus G_r}\Delta(i\mid G_r)\nonumber\\
 &\leq k\,[F(G_{r+1})-F(G_r)].
\end{align}
Iterating this residual contraction from $G_0=\varnothing$ yields
\begin{align}
 F(G_k)&\geq[1-(1-1/k)^k]F(O_k)\nonumber\\
       &\geq(1-1/e)F(O_k).
\end{align}
The first $k$ steps of a complete greedy ordering are precisely $G_k$. Thus the guarantee holds for every prefix of the same ordering, deterministically and without a union bound over $k$. This is the cardinality-constrained greedy argument of \citet{nemhauser1978analysis}, specialized to SSR's coverage objective. It makes no assumption that judge scores are calibrated to downstream reward and establishes no reward approximation factor.\hfill$\square$

\subsection{Single-Round Probability Model}
\label{app:ssr-probability}
\paragraph{Conditioning and target.}
Condition on the complete pre-race history, the finite task sets, the parent, all generated children, and the fixed RER records. Let $\mathcal C=\{H_t,H_t^{(k_1)},\ldots,H_t^{(k_M)}\}$, with $L=M+1$ candidate indices. For harness $h$ and task $x$, let $m(h,x)\in[0,1]$ be its expected reward under a fresh execution. Our fixed target is
\begin{equation}
 \mu(h)=\frac{1}{|\mathcal V|}\sum_{x\in\mathcal V}m(h,x).
 \label{eq:ssr-finite-target}
\end{equation}
This target exists even if the candidates were influenced by earlier rounds on the same validation pool. It does not identify performance on a new population of tasks.

\paragraph{Sampling conditions.}
Within the current race, the following conditions are used.
\begin{enumerate}
\item A fresh uniform permutation of $\mathcal V$ is drawn independently of fresh execution randomness. The cumulative validation sets $V_r$ are its prefixes. Unused future positions are not used to select current training tasks.
\item For each prospective stage $r\leq K$, the cumulative counts $v_r=|V_r|$ and $n_r=|P_r|$ are fixed conditional on the pre-race information, with $1\leq v_r\leq n_r$. Fixed class quotas, each capped by remaining class availability, satisfy this condition. Stopping before a stage is allowed; changing its size in response to rewards requires additional uniform control over counts.
\item Training tasks in stage $r$ are chosen using pre-race information and completed earlier stages, before any outcomes on the new batch. RER priorities and intersections of surviving candidates' evidence prefixes are allowed. Tasks are not repeated within the race.
\item For every new rollout $Y$ of a fixed harness $h$ on a task $x$, $0\leq Y\leq1$ and $\mathbb E[Y\mid\mathcal F_{\rm past},h,x]=m(h,x)$, where the past includes all information revealed before that rollout. Each active harness executes each new task $R$ times.
\end{enumerate}
The last condition is a stationary conditional-mean assumption, not merely marginal unbiasedness. Independent fresh execution randomness is a sufficient implementation. Arbitrary correlations that change the conditional mean, or a drifting environment, are not covered. Repeated rollouts do not create additional independent task draws.

\paragraph{Selection conditions.}
At intermediate stages, children are sorted by their cumulative equal-task-weight empirical scores and a prescribed number of bottom-ranked children is removed. The quota is based on the initial $M$ and at least one child remains. The parent is never removed. Ties can use diff size and then candidate index: every retained child has a score no lower than every removed child. At the last completed stage, the best surviving child replaces the parent exactly when its score is strictly higher. Consequently, the output maximizes the last-stage empirical score over the surviving children and the parent, with the parent winning ties involving itself.

\subsection{Uniform Error for Mixed Evaluation}
\label{app:ssr-uniform}
Write $T_r=P_r\setminus V_r$ and $t_r=|T_r|=n_r-v_r$. Define
\begin{align}
 \overline m_{V_r}(h)&=\frac{1}{v_r}\sum_{x\in V_r}m(h,x),\nonumber\\
 \overline m_{T_r}(h)&=\frac{1}{t_r}\sum_{x\in T_r}m(h,x)\quad(t_r>0),\nonumber\\
 \beta_r&=\max_{h\in\mathcal C}
       |\overline m_{T_r}(h)-\mu(h)|\quad(t_r>0).
 \label{eq:ssr-beta-definition}
\end{align}
Set $\beta_r=0$ when $t_r=0$. The selected training set and $\beta_r$ can be random and adaptive. The bound does not assume that these training tasks are fresh samples from the validation population. In particular, $0\leq\beta_r\leq1$, but no small value is guaranteed from the sampling rule alone.

\paragraph{Validation-task sampling.}
For each fixed harness, $\{m(h,x):x\in\mathcal V\}$ is a fixed finite population in $[0,1]$. Hoeffding's without-replacement bound gives, for $s>0$,
\begin{equation}
 \Pr\!\left(|\overline m_{V_r}(h)-\mu(h)|>s\right)
 \leq2e^{-2v_rs^2}.
 \label{eq:ssr-val-tail}
\end{equation}
See Proposition 1.2 of \citet{bardenet2015concentration}. It applies to the random validation prefix, not to RER-selected training tasks. Using $s_r=\sqrt{c_\delta/(2v_r)}$ and a union bound over at most $LK$ harness--stage pairs controls all validation deviations with failure probability at most $\delta/2$. If $v_r=|\mathcal V|$, the deviation is exactly zero, so $s_r$ may be replaced by zero at that stage.

\paragraph{Rollout noise under adaptive task choices.}
For a fixed $h$, consider its $Rn_r$ centered rewards through stage $r$,
\[
 Z_a=Y_a-m(h,x_a).
\]
Condition 4 makes these martingale differences even though later training tasks depend on earlier outcomes. Each conditional support has length one. The conditional bounded-variable moment-generating-function inequality is
\[
 \mathbb E[e^{\lambda Z_a}\mid\mathcal F_{a-1}]
 \leq e^{\lambda^2/8}.
\]
For completeness, if $g(\lambda)=\log\mathbb E[e^{\lambda Z_a}\mid\mathcal F_{a-1}]$, then $g(0)=g'(0)=0$ and $g''(\lambda)$ is the variance under exponential tilting. A variable with support length one has variance at most $1/4$, so integrating twice gives $g(\lambda)\leq\lambda^2/8$. Iterated conditioning and the exponential Markov inequality, optimized at $\lambda=4a/(Rn_r)$ for a sum deviation $a$, yield
\begin{equation}
 \Pr\left(\left|\frac{1}{Rn_r}\sum_{a=1}^{Rn_r}Z_a\right|>u\right)
 \leq2e^{-2Rn_ru^2}.
 \label{eq:ssr-rollout-tail}
\end{equation}

To make the union bound independent of survival, one may couple the process to unobserved fresh rewards for eliminated candidates on later selected tasks. These auxiliary rewards do not influence any decision and require no actual executions. They satisfy the same conditional-mean condition. A surviving harness's auxiliary score equals its observed score. Likewise, continue task selection hypothetically beyond an early budget stop using the same predictable rules. This defines all $K$ prospective stages; the theorem will only use stages actually completed. Such a continuation is possible up to the predetermined task-availability horizon.

Taking $u_r=\sqrt{c_\delta/(2Rn_r)}$ and union-bounding over $LK$ pairs controls rollout noise with failure probability at most $\delta/2$. No independence between candidates is used by the union bound; the stated conditional-mean requirement still applies to their observed rewards.

\paragraph{Combining the components.}
For every harness still active at stage $r$, the score decomposes exactly as
\begin{align}
 \widehat\mu_{h,r}-\mu(h)
 &=\left[\widehat\mu_{h,r}-\frac{1}{n_r}
                         \sum_{x\in P_r}m(h,x)\right]\nonumber\\
 &\quad+\alpha_r[\overline m_{V_r}(h)-\mu(h)]\nonumber\\
 &\quad+(1-\alpha_r)[\overline m_{T_r}(h)-\mu(h)].
 \label{eq:ssr-error-decomposition}
\end{align}
The last term is zero if $T_r$ is empty. The first bracket is centered rollout noise; the random selected-task average inside it is a predictable reward average, not necessarily a conditional expectation given the final selected set. Conditioning on that final adaptive set would generally invalidate a naive independence argument.

Combining the two concentration events proves that with probability at least $1-\delta$,
\begin{equation}
 |\widehat\mu_{h,r}-\mu(h)|\leq
 u_r+\alpha_rs_r+(1-\alpha_r)\beta_r
 =\varepsilon_r
 \label{eq:ssr-uniform-event}
\end{equation}
for every completed stage and every harness evaluated there. Because $\beta_r$ can be random, this is a pathwise, discrepancy-dependent bound. Any deterministic upper bounds $\overline\beta_r$ valid for all reachable training prefixes give a deterministic version by substitution; the unconditional choice $\overline\beta_r=1$ can be loose. The algorithm need not calculate these radii or use them as elimination thresholds.

\subsection{From Stage Errors to Selection Loss}
\label{app:ssr-elimination}
Work on the simultaneous event in Eq.~\eqref{eq:ssr-uniform-event}. Let $\mathcal S_0=\mathcal C$ and $\mathcal S_r$ contain the protected parent and children remaining after stage-$r$ elimination. Set
\[
 G_r=\max_{h\in\mathcal S_r}\mu(h).
\]
At an intermediate stage, let $b$ maximize true reward over $\mathcal S_{r-1}$. If $b$ survives, $G_r=G_{r-1}$. If it is removed, it must be a child. Any retained child $a$ has empirical score at least that of $b$, so
\begin{align}
 \mu(b)-\mu(a)
 &\leq[\mu(b)-\widehat\mu_{b,r}]
      +[\widehat\mu_{a,r}-\mu(a)]\nonumber\\
 &\leq2\varepsilon_r.
\end{align}
Therefore $G_{r-1}-G_r\leq2\varepsilon_r$. This explicitly permits the best child to be eliminated. It does not invoke the best-arm-preservation property of confidence-bound elimination \citep{evendar2006action} or assume a halving schedule \citep{karnin2013almost}.

Let $\nu\geq1$ be any last completed stage and let
\[
 \mathcal E_\nu=\{r<\nu:\text{at least one child is removed at }r\}.
\]
Telescoping only the earlier elimination stages gives
\begin{equation}
 G_0-\max_{h\in\mathcal S_{\nu-1}}\mu(h)
 \leq2\sum_{r\in\mathcal E_\nu}\varepsilon_r.
\end{equation}
If elimination also occurs at stage $\nu$ before a budget stop, a top-scoring child is still retained. The eventual parent--child comparison thus attains the empirical maximum over $\mathcal S_{\nu-1}$. For a true maximizer $b$ in that set,
\begin{align}
 \mu(b)-\mu(\widehat H)
 &\leq[\mu(b)-\widehat\mu_{b,\nu}]\nonumber\\
 &\quad+[\widehat\mu_{\widehat H,\nu}-\mu(\widehat H)]\nonumber\\
 &\leq2\varepsilon_\nu.
\end{align}
Combining the last two inequalities proves the sharper form
\begin{equation}
 \max_{h\in\mathcal C}\mu(h)-\mu(\widehat H)
 \leq2\left(\sum_{r\in\mathcal E_\nu}\varepsilon_r
                  +\varepsilon_\nu\right),
 \label{eq:ssr-sharp-selection}
\end{equation}
which implies Eq.~\eqref{eq:ssr-selection-guarantee}. No final-stage error is counted twice. Since $\widehat\mu_{\widehat H,\nu}\geq\widehat\mu_{H_t,\nu}$ and the parent remains active, the same two-error comparison gives Eq.~\eqref{eq:ssr-parent-guarantee}. If the parent is returned, its regression is exactly zero. The event is simultaneous over stages, so the result holds at the observed early stopping stage without applying a fixed-time bound after selection. If no stage completes, the method returns the parent; there is then no nontrivial selection guarantee relative to unevaluated children.\hfill$\square$

\subsection{Interpretation and Limits}
\label{app:ssr-scope}
\paragraph{A sufficient gap condition.}
Suppose $h^\star$ is the unique maximizer in $\mathcal C$ and
\[
 \Delta_{\min}=\min_{h\ne h^\star}[\mu(h^\star)-\mu(h)]>0.
\]
On the uniform event, if $2\varepsilon_r<\Delta_{\min}$ at every completed stage, then $h^\star$ strictly outranks every competing active harness empirically. It cannot be removed if it is a child, and is protected if it is the parent. Final selection returns $h^\star$. Replacing $\varepsilon_r$ with deterministic upper bounds turns this into a sufficient $1-\delta$ exact-selection condition. This is a conservative condition, not a claim of optimal sample complexity.

\paragraph{Why the discrepancy term is necessary.}
The training tasks can be directly involved in candidate generation and are selected preferentially. Their ranking need not agree with validation ranking even if rollout noise vanishes. For example, consider two children with validation gap $\mu(a)-\mu(b)=d>0$ and training mean difference $\overline m_T(a)-\overline m_T(b)=-1$. At validation weight $\alpha\in(0,1)$, their expected mixed-score difference is $\alpha d-(1-\alpha)$. It is negative whenever $d<(1-\alpha)/\alpha$. The children can therefore be misranked even with exact expected scores. This is a mathematical illustration, not experimental data. A vanishing validation-target selection bound requires a discrepancy assumption or a change to the selection objective; the proof does not silently make either change.

\paragraph{Relation to the audit and to generalization.}
The theorem compares expected rewards on a fixed finite validation pool. The racing audit instead measures a score gap against a realized full evaluation matrix; that is a related empirical diagnostic, not the same estimand. Within-round random task ordering and fresh executions justify the conditional finite-pool result even after conditioning on prior history. They do not make repeatedly reused validation tasks independent of the evolved harness for population generalization. We consequently make no multi-round or test-distribution claim from this theorem. Candidate-generation quality is also outside the selection bound.

\paragraph{More available evidence.}
For a fixed parent, optimizer mechanism, and evidence cap $k_{\max}$, define the best attainable expected reward over all eligible subsets as
\begin{align*}
 g(D)&=\mathbb E_\xi[\mu(\operatorname{Optimize}(H_t,D;\xi))],\\
 V(E)&=\max_{D\subseteq E,\,|D|\leq k_{\max}}g(D).
\end{align*}
If $E\subseteq E'$ and old configurations remain available, then $V(E')\geq V(E)$ by feasible-set inclusion. SSR explores only selected greedy prefixes, whose composition can change as the pool grows. This opportunity-set observation therefore does not establish monotonic performance for SSR, an interior optimal scale, or dominance over other evolution methods.

\section{SSR Implementation Details}
\label{app:ssr-implementation}

\paragraph{Evaluation-task construction.}
At the start of an evolution round, training tasks are partitioned into successful and failed tasks according to the current parent's collected training trajectories. These labels and the RER ordering remain fixed throughout that round. For prescribed batch sizes $b_v,b_s,b_f$, sample $b_v$ previously unused validation tasks uniformly without replacement. For each training class, first select previously unused tasks from the intersection of the surviving candidates' evidence prefixes, following the RER order. If the intersection cannot supply the class quota, continue through tasks outside the intersection in RER order. Because evidence prefixes are nested, their intersection is the shortest surviving prefix. The parent is not included in this intersection.

Exclude every task already evaluated in the current round. If a training class has fewer remaining tasks than its quota, use all remaining tasks in that class, without repetition or substitution from the other class. Likewise, the validation component is bounded by its remaining tasks. The union is shared by the parent and all active children. Equation~\ref{eq:ssr-score} weights actual tasks equally, so the effective mixture weights follow the realized task counts. Original training trajectories provide candidate-generation evidence and outcome labels; parent--child racing comparisons use the corresponding executions within the current race. Validation trajectories are not exposed to the optimizer as evidence, and the test set is never used in these decisions.

\paragraph{Integer quotas and ties.}
One deterministic integer implementation retains
\[
 \begin{gathered}
 n_r=\max\{1,M-\lfloor rM/K\rfloor\},
 \\
 r=1,\ldots,K-1,
 \end{gathered}
\]
children after each intermediate stage, with $n_0=M$. Thus stage $r$ eliminates $n_{r-1}-n_r$ children. The quotas are fixed before racing and approximate a constant quota of $M/K$ removals per stage. After the final completed stage, select the best survivor. Rank children first by cumulative score, then by ascending diff size (total added plus deleted lines relative to the parent across editable harness files), and finally by fixed candidate index. Equal RER marginal gains can likewise be resolved by fixed record index. A child tied with the parent never replaces it. These are empirical comparisons without a confidence threshold.

\paragraph{Budget boundaries and stopping.}
With active child set $\mathcal{A}$ and proposed new batch $B_r$, start the stage only if the remaining execution budget covers $R(|\mathcal{A}|+1)|B_r|$ executions. Otherwise end the race and select using its last fully completed stage. If no stage was completed, retain the parent. A full evidence refresh requires execution budget for all $N$ training executions; if that refresh cannot be completed, terminate and return the incumbent. No partial stage enters the cumulative comparison. Stop after $P$ consecutive rounds without a strict improvement, resetting the counter after an accepted update. If the parent is unchanged, reuse the collected trajectories, judge records, scores, embeddings, and ranking, but generate fresh candidates. Within-race task scores do not automatically carry into a new race.

\paragraph{Judge and optimizer interface.}
The judge is prompted to output $w_i,a_i,u_i\in[0,1]$ directly; there is no additional score normalization. It applies the same transferability criteria to successful and failed trajectories. Similarities use embeddings of the judge's trajectory summaries and extracted experience, with negative cosine values clipped to zero. The optimizer initially receives the selected experience summaries and improvement suggestions plus the parent harness. It can retrieve original trajectories for those selected records on demand, and can modify instructions, tool-related code, context management, and execution flow. Concrete model identifiers, scale choices, task quotas, repetitions, and stopping settings are reported in Appendix~\ref{app:experimental-configuration}.

\section{Experimental Configuration}
\label{app:experimental-configuration}

Table~\ref{tab:ssr-configuration} lists SSR configurations. The main experiments use Codex with GPT~5.5 high and Claude Code with Opus~4.8 high; all analyses use the Codex setting. Within each setting, the executor, optimizer, and experience judge use the same model and reasoning effort. Following Retro, the embedding model is \texttt{BAAI/bge-large-en-v1.5}, with 1{,}024-dimensional outputs \citep{pan2026retrospective}. The final SSR harnesses for SWE-Pro-V and DeepSWE under Codex are reproduced in Appendix~\ref{app:evolved-harnesses}.

\begin{table*}[!ht]
\centering
\begingroup
\footnotesize
\setlength{\tabcolsep}{3pt}
\renewcommand{\arraystretch}{1.13}
\providecolor{SSRConfigHead}{HTML}{EDF1F5}
\providecolor{SSRConfigBand}{HTML}{F3F6F8}
\newcommand{\cfgcell}[1]{\begin{tabular}[t]{@{}l@{}}#1\end{tabular}}
\begin{tabularx}{\textwidth}{@{}>{\raggedright\arraybackslash}p{0.18\textwidth}*{6}{>{\raggedright\arraybackslash}X}@{}}
\toprule
\rowcolor{SSRConfigHead}
\textbf{Parameter} & \textbf{Main} & \textbf{Task-pool growth} & \textbf{Resource growth} & \textbf{Fixed strategies} & \textbf{Racing audit} & \textbf{Scale dynamics} \\
\midrule
Benchmarks & S / D / G & S & S & S / D & S & S \\
Train / validation / test & \cfgcell{S,G: 80/20/100\\D: 50/13/50} & $0.8T$ / $0.2T$ / 100 & 80/20/100 & \cfgcell{S: 80/20/100\\D: 50/13/50} & Separate 160-task audit set & \cfgcell{(a) 160/20/100\\(b) 80/20/100} \\
\midrule
\rowcolor{SSRConfigBand}
\multicolumn{7}{@{}l}{\textbf{Candidate generation}} \\
Children $M$ & 4 & 4 & 4 & 4 & 12 & \cfgcell{(a) 6\\(b) 4} \\
Evidence scales & \cfgcell{$25/50/$\\$75/100\%$} & Main & Main & \cfgcell{25\% or 100\%\\(see text)} & Main; three children per scale & \cfgcell{(a) $10/20/40/$\\\quad$80/120/160$\\(b) Main} \\
Raw-trajectory access & \multicolumn{6}{l}{On demand, for records selected as candidate-generation evidence.} \\
Cumulative candidate cap & None & None & \cfgcell{15 / 20 / 25\\30 / 35} & None & --- & (b) None \\
\midrule
\rowcolor{SSRConfigBand}
\multicolumn{7}{@{}l}{\textbf{Racing and selection}} \\
Stages $K$ & 2 & 2 & 2 & 2 & 2 & (b) 2 \\
Validation quota $b_v$ & \cfgcell{S,G: 10 / 10\\D: 7 / 6} & \cfgcell{4 / 8 / 12\\16 / 20} & 10 & \cfgcell{S: 10 / 10\\D: 7 / 6} & --- & (b) 10 \\
Parent-success quota $b_s$ & \cfgcell{S,G: 5\\D: 3} & \cfgcell{2 / 4 / 6\\8 / 10} & 5 & \cfgcell{S: 5\\D: 3} & --- & (b) 5 \\
Parent-failure quota $b_f$ & \cfgcell{S,G: 5\\D: 3} & \cfgcell{2 / 4 / 6\\8 / 10} & 5 & \cfgcell{S: 5\\D: 3} & --- & (b) 5 \\
Rollouts per pair $R$ & 2 & 2 & 2 & 2 & 2 & (b) 2 \\
Child quotas $n_0/n_1$ & 4 / 2 & 4 / 2 & 4 / 2 & 4 / 2 & 12 / 6 & (b) 4 / 2 \\
\midrule
\rowcolor{SSRConfigBand}
\multicolumn{7}{@{}l}{\textbf{Evolution budget and stopping}} \\
Execution budget $B$ & \cfgcell{S,G: 500\\D: 300} & \cfgcell{200/400/600\\800/1{,}000} & \cfgcell{800/1{,}600\\2{,}400/3{,}200\\4{,}000} & \cfgcell{S: 500\\D: 300} & Fixed-matrix analysis & (b) 500 \\
Patience $P$ (rounds) & 2 & 2 & 2 & 2 & --- & (b) 2 \\
\bottomrule
\end{tabularx}

\vspace{4pt}
\begin{minipage}{\textwidth}
\footnotesize
\textit{Reading the table.} $T\in\{40,80,120,160,200\}$ is the total training-plus-validation pool in the task-pool analysis. Its listed batch quotas apply to each of the two stages at the corresponding $T$; main and fixed-strategy $b_v$ entries explicitly give stage~1 / stage~2. Child quotas exclude the parent. ``Main'' denotes inherited settings. ``None'' means no additional cumulative candidate-generation cap, beyond execution-budget and stopping limits. Racing and stopping entries in the last column apply only to (b); (a) directly tests individual candidates. A dash denotes a setting not tabulated for the audit. Resource-growth candidate caps are cumulative over the entire run; the corresponding MGM caps are $55/84/96/127/145$.

\end{minipage}
\endgroup

\caption{Experimental configurations. S, D, and G denote SWE-Pro-V, DeepSWE, and GAIA-2. In the scale-dynamics column, (a) denotes the parent--scale probes and (b) the final-winner scale analysis. Lists in the task-pool and resource columns follow increasing task-pool size and execution budget, respectively.}
\label{tab:ssr-configuration}
\end{table*}

\paragraph{Evidence and repetitions.}
Percentage scales round up, $k=\lceil\alpha N\rceil$: $20/40/60/80$ tasks for S/G and $13/25/38/50$ for D. \textsc{Top K Fixed} uses the same RER top 25\% for all four candidates; \textsc{Top K Resampling} draws an independent uniform 25\% subset for each; \textsc{Fixed N} uses 100\%. Budgets include training-trajectory collection and parent/child racing executions. Training evidence uses one rollout per task, refreshed after an accepted update; final testing uses three, outside the execution budget. The audit averages three task-order replays of a fixed $13\times160\times2=4{,}160$-execution matrix. Scale probes test six candidates per parent, one per scale; winner frequencies aggregate 20 independent five-round runs. Stop at execution-budget exhaustion or $P=2$ consecutive rounds without strict improvement.

\section{Variability of the Main Results}
\label{app:main-results-variance}

Every entry of Table~\ref{tab:exp-main} is the mean over three independent evolution runs of the same condition, each followed by the standard test protocol of three evaluations per test task. Table~\ref{tab:exp-main-variance} repeats these means with the standard deviation of the three run-level scores. Variances range from 0.07 to 0.96\%$^2$, so standard deviations range from 0.27 to 0.98\%. SSR's standard deviations lie between 0.27 and 0.54\%, and its differences from the strongest baseline in Table~\ref{tab:exp-main} range from 1.3\% to 3.0\%. With three runs per condition, we report this variability descriptively and do not attach significance tests to individual comparisons.

\begin{table*}[!ht]
\centering
\small
\providecolor{SSRHeader}{HTML}{EDF1F5}
\providecolor{SSRHighlight}{HTML}{DAEDF5}
\setlength{\tabcolsep}{5pt}
\renewcommand{\arraystretch}{1.07}
\begin{tabular}{lcccccc}
\toprule
\rowcolor{SSRHeader}
& \multicolumn{3}{c}{\textbf{Codex: GPT 5.5 high}} & \multicolumn{3}{c}{\textbf{Claude Code: Opus 4.8 high}} \\
\rowcolor{SSRHeader}
Method & SWE-Pro-V & DeepSWE & GAIA-2 & SWE-Pro-V & DeepSWE & GAIA-2 \\
\midrule
Initial harness & 57.7 $\pm$ 0.72 & 59.3 $\pm$ 0.54 & 29.0 $\pm$ 0.98 & 56.3 $\pm$ 0.98 & 56.7 $\pm$ 0.54 & 26.0 $\pm$ 0.54 \\
Meta-Harness & 59.7 $\pm$ 0.82 & 60.7 $\pm$ 0.54 & 31.7 $\pm$ 0.94 & 60.0 $\pm$ 0.47 & 58.0 $\pm$ 0.94 & 29.3 $\pm$ 0.82 \\
HGM & 58.7 $\pm$ 0.98 & 62.7 $\pm$ 0.94 & 32.3 $\pm$ 0.94 & 60.3 $\pm$ 0.72 & 58.7 $\pm$ 0.94 & 29.3 $\pm$ 0.82 \\
MGM & 60.0 $\pm$ 0.94 & 63.3 $\pm$ 0.54 & 32.3 $\pm$ 0.72 & 61.0 $\pm$ 0.47 & 60.7 $\pm$ 0.94 & 29.7 $\pm$ 0.54 \\
AHE & 60.7 $\pm$ 0.54 & 65.3 $\pm$ 0.54 & 36.0 $\pm$ 0.82 & 62.0 $\pm$ 0.98 & 63.3 $\pm$ 0.54 & 33.7 $\pm$ 0.47 \\
Retro & 62.7 $\pm$ 0.47 & 66.7 $\pm$ 0.94 & 36.3 $\pm$ 0.98 & 61.3 $\pm$ 0.54 & 64.0 $\pm$ 0.98 & 34.3 $\pm$ 0.72 \\
\midrule
\rowcolor{SSRHighlight}
\textbf{SSR (ours)} & \textbf{65.7} $\pm$ 0.54 & \textbf{68.0} $\pm$ 0.54 & \textbf{38.0} $\pm$ 0.27 & \textbf{63.3} $\pm$ 0.47 & \textbf{65.3} $\pm$ 0.54 & \textbf{36.0} $\pm$ 0.47 \\
\bottomrule
\end{tabular}

\caption{Final harness performance as mean $\pm$ standard deviation (Avg Pass@1, \%) over three independent evolution runs. Means are those of Table~\ref{tab:exp-main}; bold marks the best mean in each setting.}
\label{tab:exp-main-variance}
\end{table*}

\section{Evidence Scale across Parents and Evolution Rounds}
\label{sec:exp-scale-dynamics}

\begin{figure}[!t]
\centering
\includegraphics[width=\columnwidth]{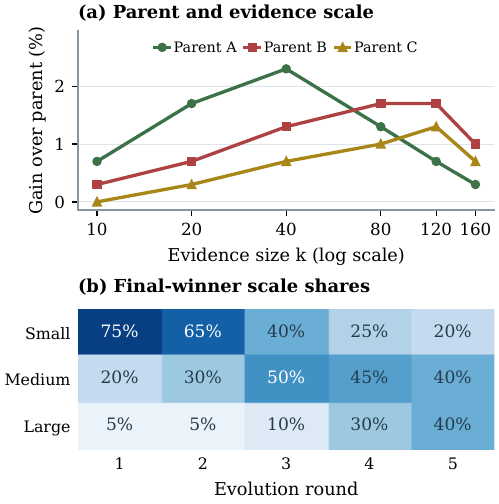}
\caption{Evidence scales across parents and rounds. (a) Single-candidate gains (one per scale, no intervals) for parents from different independent runs. (b) Final-winner scale shares over 20 runs; each column sums to 100\%. These are winner frequencies, not survival probabilities.}
\label{fig:exp-scale-dynamics}
\end{figure}

\paragraph{Scale and parent state.}
We randomly sample A/B/C from rounds 1/2/3 of different independent evolution runs. With an expanded training pool, each parent generates one candidate per RER prefix at $k\in\{10,20,40,80,120,160\}$. All 18 candidates and their parents use the main test protocol.

The largest observed gains occur at $k=40$ for A, $k=80$ or $120$ for B, and $k=120$ for C (Figure~\ref{fig:exp-scale-dynamics}a). No scale maximizes all three curves. These single candidates illustrate variation across snapshots, rather than estimating mean optimal scales or establishing an evolution-stage effect.

\paragraph{Winning scales across rounds.}
Across 20 runs of five rounds, small-scale winners decline from 75\% to 20\%, while large-scale winners rise from 5\% to 40\% (Figure~\ref{fig:exp-scale-dynamics}b). The best scale thus varies across parents, and later rounds more often select winners generated from broader evidence.

\section{More Related Work}
\label{app:more-related-work}

\paragraph{Racing and successive halving.}
Selecting among costly-to-evaluate candidates under a fixed budget is a classical problem. Racing algorithms evaluate all survivors on a shared, growing set of instances and eliminate candidates by statistical tests \citep{maron1994hoeffding,birattari2002racing,lopezibanez2016irace}. Successive halving and Hyperband instead remove a fixed fraction of candidates at each stage \citep{karnin2013almost,jamieson2016nonstochastic,li2018hyperband}, and best-arm identification analyses bound the resulting selection error \citep{evendar2006action,audibert2010best}. Incumbent-based intensification in algorithm configuration compares challengers with the current best on common instances and replaces it only on strict improvement \citep{hutter2009paramils,hutter2011smac}. TRIPLE applies successive halving to select among a fixed pool of prompts \citep{shi2024triple}. SSR's candidate racing combines fixed-quota elimination with a protected parent. It differs in what is raced and on what: the candidates are generated from nested evidence prefixes of one parent, evaluation batches mix held-out validation tasks with training tasks chosen from the survivors' evidence, and the pool is regenerated as the parent evolves.

\clearpage\onecolumn
\section{Evolved Harnesses}
\label{app:evolved-harnesses}

This appendix reproduces the final harnesses evolved by SSR in the Codex setting for SWE-Pro-V and DeepSWE, the harnesses behind the SSR entries of 65.7\% and 68.0\% in Table~\ref{tab:exp-main}. Each harness consists of Markdown instruction files and one shell script, which SSR added to the platform's default tools and runtime (Section~\ref{sec:exp-setup}). Files are listed verbatim under their file names; wrapped lines are marked with $\hookrightarrow$. The instruction files refer to the template and the script by the paths under which the harness is mounted, which are noted with each file.

\subsection{SWE-Pro-V Harness}
\label{app:harness-swe-pro-v}

\paragraph{\texttt{Readme.md}.} Entry instructions.
\begin{lstlisting}[style=harnessfile]
# Solver Harness

Use this harness as a small operating procedure, not as a replacement for reading the repository. The recurring failures in prior runs came from incomplete surface coverage, weak contract tests, unfinished verification, missed toolchains, and unsafe temporary files.

Start with [workflows/contract-driven-solve.md](workflows/contract-driven-solve.md). Keep a short matrix from [templates/contract-matrix.md](templates/contract-matrix.md) for broad tasks.

## Default Workflow

1. Convert the prompt into concrete contracts: public interfaces, behavior, edge cases, defaults, persistence shape, error text, exit codes, and compatibility notes.
2. Search for every affected surface before editing: direct callers, old field/type users, alternate constructors, proxy/non-interactive paths, tests, serialization, config, CLI flags, and hidden fallback paths.
3. Implement through the repository's existing abstractions. Avoid adding evaluator-only helpers or test shims unless the prompt explicitly requires production code.
4. Add focused regression coverage for the contract, especially boundary states and exact serialized/request payloads.
5. Finish only after verification commands have exited and their status is known. Use `harness/bin/solve-verify` to discover tools, run focused checks, and flag unsafe `/tmp` additions.

## Runtime Contract

- Persistent solve state belongs under `$RHO_SCRATCH_DIR`, which points at `task/.harness-state/`.
- Do not write matrices, ledgers, reports, build outputs, or notes to a fixed `/tmp/...` path.
- Disposable temporary files must use `mktemp` with a unique path and must be removed by the creating process.
- Reading shared tools or using concurrency-safe tool-managed caches under `/tmp` is allowed.

## Useful Commands

From a task repository root:

```bash
harness/bin/solve-verify
harness/bin/solve-verify -- go test ./changed/package
harness/bin/solve-verify -- yarn lint:types
```

The helper also looks for Go tools in common locations such as `/usr/local/go/bin` when they are not on `PATH`.
\end{lstlisting}

\paragraph{\texttt{workflow.md}.} Referred to from \texttt{Readme.md} as \texttt{workflows/contract-driven-solve.md}.
\begin{lstlisting}[style=harnessfile]
# Contract-Driven Solve Workflow

Use this for tasks with many requirements, shared interfaces, parsers, CLIs, persistence, async behavior, or UI state.

## 1. Build the Contract Matrix

Before editing, copy `templates/contract-matrix.md` into `$RHO_SCRATCH_DIR` or keep the same information in your notes. For each requirement, record:

- the exact public interface or behavior requested
- the implementation files and call paths that must change
- the edge cases that would prove it
- the command or test that verifies it

Do not treat a compile check or a smoke run as enough when the prompt specifies serialized payloads, exact defaults, filtering, status transitions, boundary UI behavior, or error text.

## 2. Search the Whole Affected Surface

After identifying the main change, search for bypass paths:

- old type/field/function names and direct calls to the old behavior
- alternate constructors and config defaulting paths
- CLI flags, config parsers, web/API entry points, non-interactive flows, proxy/recording paths, and background workers
- interface implementations, mocks, wrappers, generated clients, and tests
- persistence keys, secondary indexes, request payload builders, and response hydration helpers

When semantics change from boolean to enum, sync to async, static address to runtime address, literal parsing to structured parsing, or derived state to persisted state, every old direct path is suspect until searched.

## 3. Write Acceptance-Style Regressions

Prefer small tests that pin the public contract over large broad tests that only compile. Good regression targets from prior failures:

- parser/adapter tasks: historical and current input shapes, nil versus empty slices, deterministic ordering, identifier preference, exact output JSON/request body, filtering and exit-code behavior
- CLI/testability tasks: option injection, no process exit on parser or handler errors, error preservation, and dynamic `:0` listener address propagation
- async/event tasks: non-blocking overflow, backoff/drop counters, close behavior, all emission paths routed through the new abstraction
- state-transition tasks: no resources, one resource, newest resource removed while older remain, final resource removed, monotonic timestamps in UTC
- UI/control tasks: initial zero state, start, middle, exact end, out-of-range/invalid input, state and position observers staying in sync
- import/packaging tasks: end-to-end payload contents, redirects, missing package `__init__.py`, relative imports from package initializers, and precise error messages
- storage lifecycle tasks: create, list order, get single/array/empty, update preserving unrelated fields, delete cleanup from every index, null last-seen values

For ambiguous prompt text, obey explicit local compatibility or repair notes over generic requirements. Do not add helper packages, `export_test.go` shims, or fake test environments that the prompt says hidden tests will supply.

## 4. Verify Deliberately

Find available tools before giving up. In Go repositories, check both `go` on `PATH` and `/usr/local/go/bin/go` plus `/usr/local/go/bin/gofmt`.

Run the smallest tests that prove the changed contract, then run adjacent package checks when interfaces or shared behavior changed. Record every command's exit status; if a command is still running, failed, or was skipped, say so in the final answer.

Before finishing, run:

```bash
harness/bin/solve-verify
```

It reports common tool locations and scans the current diff for added fixed `/tmp` paths.
\end{lstlisting}

\paragraph{\texttt{template.md}.} Referred to as \texttt{templates/contract-matrix.md}.
\begin{lstlisting}[style=harnessfile]
# Contract Matrix

Store task-specific copies under `$RHO_SCRATCH_DIR`, not under `harness/` and not at a fixed `/tmp` path.

| Requirement / edge case | Implementation surface | Bypass paths searched | Regression or command | Result |
| --- | --- | --- | --- | --- |
|  |  |  |  |  |
\end{lstlisting}

\paragraph{\texttt{solve-verify.sh}.} Invoked as \texttt{harness/bin/solve-verify}.
\begin{lstlisting}[style=harnessfile]
#!/usr/bin/env bash
set -euo pipefail

usage() {
  cat <<'EOF'
Usage:
  harness/bin/solve-verify
  harness/bin/solve-verify -- <verification command> [args...]

With no command, prints discovered tool paths and scans the current git diff for
added fixed /tmp paths. With a command, also runs it and records the exit status.
Persistent logs are written only under $RHO_SCRATCH_DIR when it is set.
EOF
}

find_tool() {
  local name="$1"
  local path

  if path="$(command -v "$name" 2>/dev/null)"; then
    printf '%s=%s\n' "$name" "$path"
    return 0
  fi

  for path in \
    "/usr/local/go/bin/$name" \
    "/usr/local/bin/$name" \
    "/opt/homebrew/bin/$name" \
    "/usr/bin/$name" \
    "/bin/$name"; do
    if [[ -x "$path" ]]; then
      printf '%s=%s\n' "$name" "$path"
      return 0
    fi
  done

  printf '%s=<not found>\n' "$name"
  return 1
}

print_tools() {
  local name
  for name in rg go gofmt node npm yarn python python3 pytest; do
    find_tool "$name" || true
  done
}

make_temp_dir() {
  mktemp -d "${TMPDIR:-/tmp}/rho-solve-verify.XXXXXX"
}

scan_fixed_tmp_diff() {
  if ! git -c safe.directory='*' rev-parse --show-toplevel >/dev/null 2>&1; then
    echo "git_diff_tmp_scan=skipped (not a git worktree)"
    return 0
  fi

  local tmpdir diff_file
  tmpdir="$(make_temp_dir)"
  trap 'rm -rf "$tmpdir"' RETURN
  diff_file="$tmpdir/diff.patch"

  if ! git -c safe.directory='*' diff -- . ':(exclude)harness/**' >"$diff_file" 2>/dev/null; then
    echo "git_diff_tmp_scan=skipped (git diff failed)"
    return 0
  fi

  if grep -E '^\+[^+].*/tmp/[A-Za-z0-9._-]+' "$diff_file" >/dev/null; then
    echo "git_diff_tmp_scan=warning"
    echo "Added lines mention fixed /tmp paths. Use \$RHO_SCRATCH_DIR for persistent state or mktemp plus cleanup for disposable files:"
    grep -n -E '^\+[^+].*/tmp/[A-Za-z0-9._-]+' "$diff_file" | sed -n '1,20p'
  else
    echo "git_diff_tmp_scan=ok"
  fi
}

run_command() {
  local log_dir log_file status

  if [[ -n "${RHO_SCRATCH_DIR:-}" ]]; then
    mkdir -p "$RHO_SCRATCH_DIR"
    log_dir="$RHO_SCRATCH_DIR"
  else
    log_dir="$(make_temp_dir)"
    trap 'rm -rf "$log_dir"' EXIT
    echo "RHO_SCRATCH_DIR is not set; command log will be disposable."
  fi

  log_file="$log_dir/verify-$(date -u +%Y%m%dT%H%M%SZ)-$$.log"
  echo "verify_log=$log_file"
  echo "verify_command=$*" | tee "$log_file"

  set +e
  "$@" 2>&1 | tee -a "$log_file"
  status=${PIPESTATUS[0]}
  set -e

  echo "verify_exit_status=$status" | tee -a "$log_file"
  return "$status"
}

if [[ "${1:-}" == "-h" || "${1:-}" == "--help" ]]; then
  usage
  exit 0
fi

print_tools
scan_fixed_tmp_diff

if [[ "${1:-}" == "--" ]]; then
  shift
  if [[ "$#" -eq 0 ]]; then
    usage >&2
    exit 2
  fi
  run_command "$@"
fi
\end{lstlisting}

\subsection{DeepSWE Harness}
\label{app:harness-deepswe}

\paragraph{\texttt{Readme.md}.} Entry instructions.
\begin{lstlisting}[style=harnessfile]
# DeepSWE solve harness

Use this harness to turn a task prompt into observable acceptance checks before
the implementation starts. A green pre-existing suite is regression evidence;
it is not evidence that a new feature satisfies its contract.

## Non-negotiable runtime rules

- Multiple workers may share the host `/tmp`. Never keep task-specific mutable
  state at a fixed global `/tmp/...` path.
- Put matrices, notes, logs, patches, reports, and other persistent solve state
  under `$RHO_SCRATCH_DIR` (the runner creates `task/.harness-state/` and exports
  its absolute path).
- A truly disposable file or directory may be made with `mktemp`; the process
  that creates it must remove it.
- Reading shared tools or using concurrency-safe tool-managed caches under
  `/tmp` is fine.
- In container tasks, repository changes and commits belong in `/app`, not in
  the host workspace. Treat `/host-ws` and `$RHO_SCRATCH_DIR` as staging and
  evidence locations only.

Initialize the acceptance ledger without overwriting an existing one:

```sh
harness/bin/init-acceptance-matrix
```

If the harness is mounted elsewhere, invoke that script by its discovered
absolute path. It writes only to `$RHO_SCRATCH_DIR`.

For a container repository that cannot be edited directly, keep a candidate
patch in `$RHO_SCRATCH_DIR/change.patch`, check it before mutation, then apply
the same bytes:

```sh
docker exec -i "$container" sh -lc 'cd /app && git apply --check -' \
  <"$RHO_SCRATCH_DIR/change.patch"
docker exec -i "$container" sh -lc 'cd /app && git apply -' \
  <"$RHO_SCRATCH_DIR/change.patch"
```

Prefer several small checked patches over one large shell-quoted rewrite.

## The solve loop

### 1. Establish the real repository contract

Before editing:

1. Read repository instructions and the prompt completely.
2. Confirm the repository root, current revision, available toolchain, and
   test/build conventions. Inspect nearby implementation and tests.
3. If requested, create a branch from `main` before changes. Configure Git
   identity repository-locally only if a commit requires it.
4. Record prohibited tooling, offline constraints, exact public names, exact
   errors, submission/commit requirements, and supported modes in the ledger.

Do not download a generator or replace checked-in generated artifacts when the
task requires the checkout's existing toolchain. Do not broaden the task with
unrelated cleanup, completion, documentation, or sorting changes.

### 2. Derive acceptance rows from prose

Create one row for every independently observable clause. Pay special attention
to words such as **all**, **every**, **only**, **never**, **exactly**, **same**,
**before**, **after**, **until**, **even on failure**, **default**, and **must**.
Classify each row:

- `P` positive behavior
- `N` rejection, error, or prohibited behavior
- `B` boundary or unrestricted input domain
- `X` exact API, schema, representation, ordering, or message
- `C` composition with another feature, mode, or existing API
- `L` lifecycle, state transition, concurrency, or cleanup
- `R` backward compatibility and unchanged behavior

For each row write the public entrypoint, setup, action, exact observable
result, and oracle. A helper-level assertion is insufficient when the contract
is visible through a CLI, transport, renderer, compiler, serializer, or other
public path.

### 3. Build the smallest high-risk interaction matrix

Do not blindly test a full Cartesian product. Use pairwise cases plus cases at
representation and lifecycle boundaries. At minimum cover:

- every explicitly accepted syntax/value/API variant, including alternate
  grammar branches and aliases;
- every execution mode/backend/public access path named by the prompt;
- every state-changing builder operation followed by every terminal mode it
  can affect;
- empty, zero, nil/null, missing, duplicate, maximum/exact-limit,
  just-over-limit, whitespace, Unicode, special property names, and invalid
  values where legal;
- first use, repeated use, reset/reuse, removal, failure, and use after
  serialization or reconstruction;
- operation-order reversals: configure before/after registration, abort before
  start/during pending work/after completion, and callbacks on every exit path;
- disjoint concurrent inputs, out-of-order completion, one stalled worker,
  cleanup races, and bounded-retention pressure when concurrency is required;
- a new feature composed with the closest existing API, rather than only used
  alone;
- processing exactly once versus repeated processing, especially pipelines
  that coalesce, render, merge, or replay in multiple stages.

Turn a category definition into both qualifying and non-qualifying cases. Turn
an unconditional statement into an empty-input case. Turn an ordering statement
into multiple items whose insertion, lexical, priority, and completion orders
differ.

### 4. Choose semantic, implementation-independent oracles

Test what the caller observes, not the trigger or representation chosen by the
implementation.

- Serialization/output: assert exact keys and types, empty arrays versus null,
  ordering, section counts, comments, whitespace, and final newlines. Then parse
  or decode and assert semantics.
- Transformations/optimizers: compare before/after behavior over the supported
  domain. Include a preservation case and a locality counterexample proving
  unrelated input still changes when allowed.
- Parsers: cover every syntax container, contextual fallback, exact diagnostic,
  malformed delimiter/quote case, and compatibility when the feature is off.
- Stateful systems: assert the transition, current snapshot, side effects,
  cleanup, and subsequent operation—not merely stored fields.
- Caches/concurrency: vary worker inputs. Assert preserved union and actual reuse
  or work counts, not merely valid files/checksums.
- Resource bounds: instrument every queue, collector, formatter, and replay
  stage. Compare default-concurrency output byte-for-byte with the baseline.
- Async/UI: drive real event paths and passive state changes. Do not inject an
  implementation-specific event merely to make the test pass.
- Binary formats: decode and assert every requested field with known nonzero
  values; structural validity alone is too weak.
- Type-level APIs: add compile-pass and compile-fail checks as well as runtime
  tests.

Keep focused acceptance tests in the repository when its conventions allow.
Never weaken, narrow, or delete a regression just because it exposes an
architectural problem.

### 5. Trace the complete data and state flow

Before settling on an architecture, trace the feature from public input to the
final consumer. List every parsing, normalization, caching, persistence,
rendering, reconstruction, and cleanup stage. This catches repeated processing
and metadata lost at boundaries.

Check these questions explicitly:

- What is the true identity key? If behavior depends on variables, candidate,
  target, path, scope, or generation, a name-only cache is probably too coarse.
- Who owns mutable state, and who clears it on removal, reset, close, error, or
  global-state replacement?
- Does the implementation need to distinguish absence from a legal zero/empty
  value? Use an explicit presence bit rather than a value sentinel.
- Does a reused legacy helper implement the prompt's exact category boundary?
- Is an immutable API's returned value retained?
- Can callbacks, timing, metrics, or cleanup be skipped by an early return?
- Does stored metadata survive every persistence/reconstruction route?
- Can shared source input be mutated and leak into siblings or later passes?

### 6. Validate in a cost-aware ladder

Run checks in this order and record the command and exit status in the ledger:

1. smallest compile/type/format check for touched code;
2. focused acceptance tests, including negative and interaction rows;
3. affected package/module tests;
4. public CLI/API/transport smoke test using the built artifact;
5. broader regression suite, lint, vet, race, or build as appropriate;
6. final diff, exact API/exports/docs/generated-file audit, and commit checks.

When a broad command is blocked by an offline dependency or unrelated fixture,
run the narrow suites that exercise the changed code and report the blocker
precisely. Do not claim later commands in a shell chain ran if an earlier command
failed. A failure found by a strong lifecycle test must be fixed or explicitly
reported, never replaced with a weaker scenario.

Avoid waste: inspect before patching, use small verified patches, stop repeating
unchanged failing commands, and do not install unavailable tools merely to make
one optional check green.

### 7. Submission gate

Before the final answer:

- Re-read every prompt clause and mark each ledger row `pass`, `blocked`, or
  `unresolved`, with evidence.
- Run `git diff --check` and inspect the final diff for accidental or generated
  changes.
- Confirm focused tests are included when appropriate.
- Commit every intended repository change if the task grades committed work.
- Confirm the branch/base requirement and inspect `git status --short` and the
  final commit. Do not call a dirty or partially committed tree clean.
- State exactly what passed, what was blocked, and any remaining contract risk.
\end{lstlisting}

\paragraph{\texttt{template.md}.} Referred to as \texttt{templates/acceptance-matrix.md}; copied into the scratch directory by the script below.
\begin{lstlisting}[style=harnessfile]
# Acceptance matrix

Task:
Repository / revision:
Branch requirement:
Toolchain and offline constraints:
Exact submission requirement:

## Public contract inventory

List exact names, signatures, exports, errors, schemas, ordering, formatting,
defaults, feature gates, and behavior that must remain unchanged.

| ID | Kind | Prompt clause | Public path | Setup and action | Exact observable / oracle | Test | Status |
|---|---|---|---|---|---|---|---|
| A01 | P | | | | | | todo |
| A02 | N | | | | | | todo |
| A03 | X | | | | | | todo |

Kinds: `P` positive, `N` negative, `B` boundary, `X` exact contract,
`C` composition, `L` lifecycle/concurrency, `R` compatibility.

## Dimension inventory

Fill only relevant dimensions, but enumerate every explicit variant.

| Dimension | Values required by the prompt | High-risk boundary/contrast |
|---|---|---|
| public execution path | | |
| mode / backend / feature gate | | |
| syntax or value category | | |
| structure / nesting / representation | | |
| state / lifecycle phase | | |
| failure / cancellation phase | | |
| concurrency / ordering | | |
| empty / zero / nil / missing | | |
| exact output / schema / types | | |
| existing API composition | | |

## High-risk pairwise cases

Record deliberate combinations. Avoid a full Cartesian product unless the
state space is small.

| Case | Dimensions crossed | Why this boundary is risky | Expected observation | Test/status |
|---|---|---|---|---|
| C01 | | | | todo |

## State and ownership model

True identity key:
State owner:
Presence versus zero/empty representation:
Reset/removal/close/error cleanup:
Persistence/reconstruction boundaries:
Repeated-processing or idempotence risk:

| From state | Operation / event | Failure or ordering variant | To state | Side effects and cleanup |
|---|---|---|---|---|
| | | | | |

## End-to-end data flow

Public input -> parsing -> normalization -> core operation -> persistence/cache
-> reconstruction/rendering -> public output

At each arrow note metadata that must survive and whether the stage can run more
than once.

## Validation evidence

Record commands separately so one failure cannot mask whether another ran.

| Command | Revision | Scope / acceptance IDs | Exit status | Notes / blocker |
|---|---|---|---|---|
| | | | | |

## Final gate

- [ ] Every prompt clause maps to an acceptance row.
- [ ] Positive, negative, boundary, exact-contract, composition, and lifecycle
      cases relevant to this task are covered.
- [ ] Public entrypoints and semantic outcomes are tested, not only helpers.
- [ ] Exact schemas/types/order/whitespace/errors are asserted where specified.
- [ ] Focused tests remain in the repository when appropriate.
- [ ] No failing regression was weakened or discarded.
- [ ] Broad validation blockers are recorded precisely.
- [ ] Final diff and `git diff --check` inspected.
- [ ] Intended changes committed when required; final status inspected.
- [ ] Persistent state is under `$RHO_SCRATCH_DIR`; disposable `mktemp` paths
      were cleaned.

\end{lstlisting}

\paragraph{\texttt{solve-verify.sh}.} Invoked as \texttt{harness/bin/init-acceptance-matrix}.
\begin{lstlisting}[style=harnessfile]
#!/bin/sh
set -eu

if [ -z "${RHO_SCRATCH_DIR:-}" ]; then
  echo 'RHO_SCRATCH_DIR is required; the runner must provide task/.harness-state' >&2
  exit 2
fi

case "$RHO_SCRATCH_DIR" in
  /*) ;;
  *)
    echo 'RHO_SCRATCH_DIR must be an absolute path' >&2
    exit 2
    ;;
esac

script_dir=$(CDPATH= cd -- "$(dirname -- "$0")" && pwd)
template="$script_dir/../templates/acceptance-matrix.md"
destination="$RHO_SCRATCH_DIR/acceptance-matrix.md"

mkdir -p -- "$RHO_SCRATCH_DIR"

if [ -e "$destination" ]; then
  printf '%s\n' "$destination"
  exit 0
fi

cp -- "$template" "$destination"
printf '%s\n' "$destination"
\end{lstlisting}

\end{document}